\documentclass{article}
\usepackage{spconf,amsmath,graphicx}
\usepackage{multirow}

\title{MULTI-VIEW FRAME RECONSTRUCTION WITH CONDITIONAL GAN}
\name{Tahmida Mahmud \qquad Mohammad Billah \qquad Amit K. Roy-Chowdhury}
\address{University of California, Riverside}
\begin{document}
%
\maketitle
\begin{abstract}
Multi-view frame reconstruction is an important problem particularly when multiple frames are missing and past and future frames within the camera are far apart from the missing ones. Realistic coherent frames can still be reconstructed using corresponding frames from other overlapping cameras. We propose an adversarial approach to learn the spatio-temporal representation of the missing frame using conditional Generative Adversarial Network (cGAN). The conditional input to each cGAN is the preceding or following frames within the camera or the corresponding frames in other overlapping cameras, all of which are merged together using a weighted average. Representations learned from frames within the camera are given more weight compared to the ones learned from other cameras when they are close to the missing frames and vice versa. Experiments on two challenging datasets demonstrate that our framework produces comparable results with the state-of-the-art reconstruction method in a single camera and achieves promising performance in multi-camera scenario.
\end{abstract}
\section{Introduction}
Looking at a video sequence with one or more missing frames, how do we infer about what happened in the missing portion? We have never visualized that missing frame. Instead we have a knowledge of the spatio-temporal context of the video to reason about a potential unknown scenario. This spatio-temporal context from the adjacent frames within the camera and the corresponding frames from other overlapping cameras is key to solving an important problem in automated video analysis- frame reconstruction - which  is the task of reconstructing one or more missing frames in videos. Frame reconstruction is critical in applications like retrieving missing frames in surveillance videos, anomaly detection, data compression, video editing, video post-processing, animation, spoofing and so on. Although there have been works on frame reconstruction in a single camera setting \cite{sun2018temporally, chen2017long, jiang2017super} to the best of our knowledge, ours is the first work to solve it in a multi-camera scenario. 

\begin{figure*}[t]
	\centering
	\includegraphics[width=0.85\linewidth] {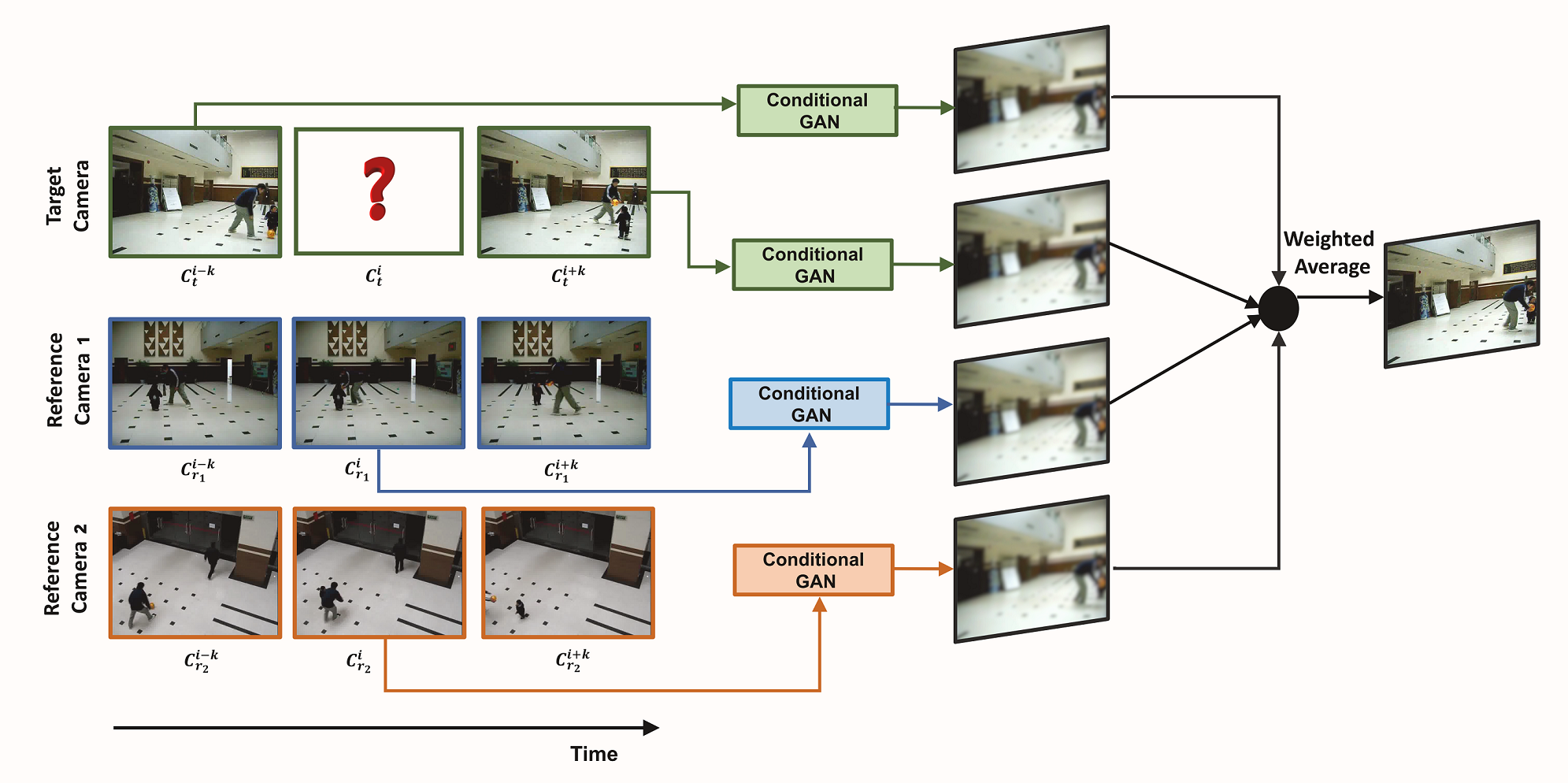}
	\caption{An example case when there are $3$ cameras and the  $i^{th}$ frame, $C_t^i$ is missing from target camera $1$ of Office Lobby Dataset \cite{fu2010multi}. We want to generate the missing frame using four available frames ($i^{th}$ frames from reference camera $2$ and $3$, $C_{r_2}^i$ and $C_{r_3}^i$ respectively, and $(i-k)^{th}$ and $(i+k)^{th}$ frames from target camera $1$, $C_t^{i-k}$ and $C_t^{i+k}$ respectively). Here, $k$ can be any arbitary number.}
	\vspace{-3mm}
	\label{fig:fig2}
\end{figure*}

\noindent {\bf Overview of Our Approach.}
We present an adversarial approach to learn a joint spatio-temporal representation of the missing frame in a multi-camera scenario. 
First, we learn the possible representations of the missing frame conditioned on the preceding and following frames within the camera as well as on the corresponding frames in other overlapping cameras using conditional Generative Adversarial Network (cGAN) \cite{mirza2014conditional} similar to the one used in \cite{isola2017image}. Then all of these representations are merged together using a weighted average where the weights are chosen as follows: representations learned from frames within the camera are given more weight when they are close to the missing frame and representations learned from frames in other overlapping cameras are given more weight when the available intra-camera frames are far apart. Overview of our proposed framework is illustrated in Fig.~\ref{fig:fig2}. The {\bf main contributions} of our work are: 
\begin{enumerate} 
	\item{We tackle a novel problem of frame reconstruction in multi-camera scenario.} 
	\item{We perform extensive experiments on a challenging multi-camera video dataset to show the effectiveness of our method.}
	\item{We perform extensive experiments on a single-camera video dataset to provide quantitative comparison of our proposed method with others in the literature.}
\end{enumerate}

\section{Related Works} \label{section:related}
Our work is related to video inpainting, frame interpolation, video prediction, frame reconstruction, and generative adversarial networks. There are important differences between frame reconstruction and the problems of video inpainting or frame interpolation. Some spatial information is available in inpainting since the missing portions are assumed to be localized to small spatio-temporal regions. Interpolation cannot reconstruct multiple missing frames as it requires the adjacent (maximum $0.05$ seconds apart \cite{sun2018temporally}) frames as inputs. In video prediction, the goal is to predict the most probable future frames from a sequence of past observations.

There are patch-based approaches \cite{newson2014video}, probabilistic model based approaches \cite{ebdelli2015video} and methods handling background and foreground separately \cite{patwardhan2007video, hung2017exemplar} for \emph{video inpainting}. For \emph{frame interpolation}, there are approaches \cite{chen2011image} using dense optical flow field, phase-based method  \cite{meyer2015phase}, deep learning approaches \cite{niklaus2017video2, liu2017video, zhou2016view} and works on long term interpolation \cite{chen2017long, jiang2017super}. There are sequence-to-sequence learning-based approaches \cite{ranzato2014video, srivastava2015unsupervised}, predictive coding network \cite{lotter2016deep}, convolutional LSTM \cite{villegas2017decomposing}, deep regression network \cite{vondrick2016anticipating} for \emph{video prediction}. The recent state-of-the-art work on \emph{frame reconstruction} within a single camera \cite{sun2018temporally} uses an LSTM-based interpolation network. However, to the best of our knowledge, there is no work performing frame reconstruction in a multi-camera scenario. This is important when adjacent available frames within the camera are far apart and frames from other corresponding overlapping views can be useful. Recently, \emph{Generative Adversarial Networks} \cite{goodfellow2014generative} have become popular to solve challenging computer vision problems like text-to-image synthesis \cite{reed2016generative}, frame interpolation \cite{van2017frame} and so on. \cite{isola2017image} has shown outstanding performance in conditional transfer of pixel-level knowledge. In this work, we seek to leverage GANs for the multi-camera reconstruction problem.

\begin{figure*}[t]
	\centering
	\includegraphics[width=0.75\linewidth] {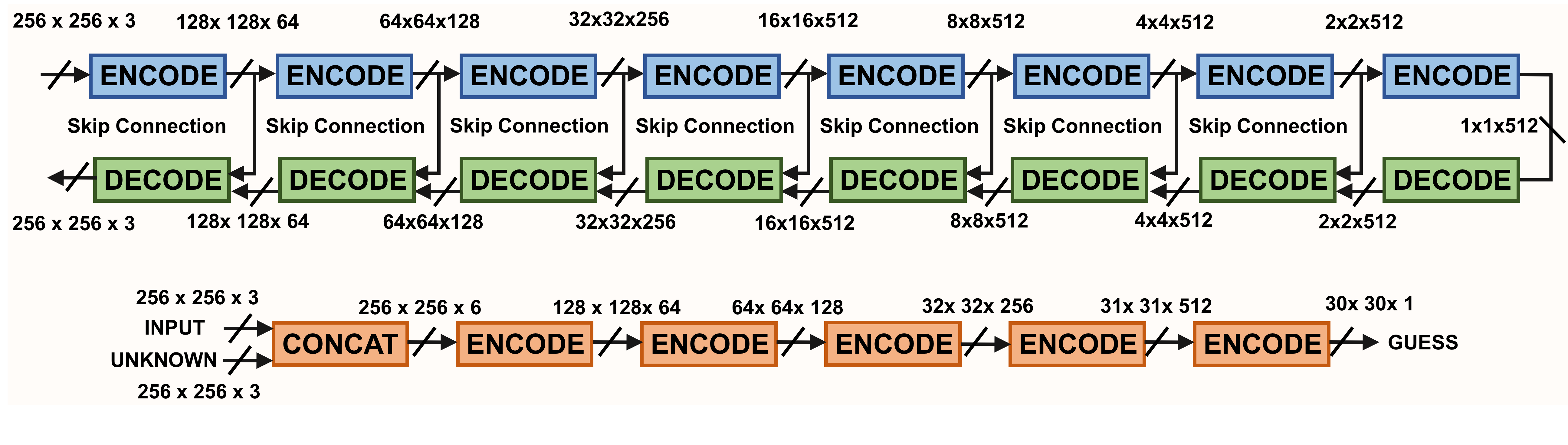}
	\caption{Proposed architecture for the generator (top) and the discriminator (bottom) \cite{isola2017image}. The pixel values in the $30 \times 30$ output show how realistic that section of the unknown image is.}
	\vspace{-3mm}
	\label{fig:fig3}
\end{figure*}

\section{Methodology} \label{section:Method}
We would refer the camera with the missing frames as the target camera and other cameras as the reference cameras. 
\subsection{Network Architecture}
Similar to general GAN, conditional GAN has a generator and a discriminator. Both of our generator and discriminator have the same architectures used in \cite{radford2015unsupervised}. We use the conditional GAN to do a mapping between inter-camera or intra-camera frames. These frames share an underlying structure  i.e., they share some common low-level information which we want to transfer across the network. Previous image translation problems used an encoder-decoder network \cite{hinton2006reducing} where the input was downsampled after being passed through a number of layers and then upsampled using a reverse process when a bottleneck layer was reached \cite{isola2017image}. We use a ``U-Net"-based architecture of the generator adding skip connection between each layer to overcome the bottleneck problem as the skip connections directly connect encoder layers to decoder layers. L1 loss efficiently captures the low frequency components of images. But using only L1 loss in the objective function for image mapping generates blurry results. We are using a combination of L1 loss and adversarial loss in the objective function. So we aim to use a discriminator efficient in modeling the high frequency components of images. We use the PatchGAN \cite{isola2017image} to focus on the structure at local image patches. The discriminator tries to differentiate between the generated and the actual missing frames at patch-level and runs convolutationally across the image to generate an averaged output. So, in this way, the image is modeled as a  Markov random field assuming that the pixels separated by more than one patch diameter are independent. The high level network architectures for the generator and discriminator are shown in Fig.~\ref{fig:fig3}.

\subsection{Model Training and Inference}\label{section:train}
\begin{figure}[h!]
	\centering
	\includegraphics[width=0.75\linewidth] {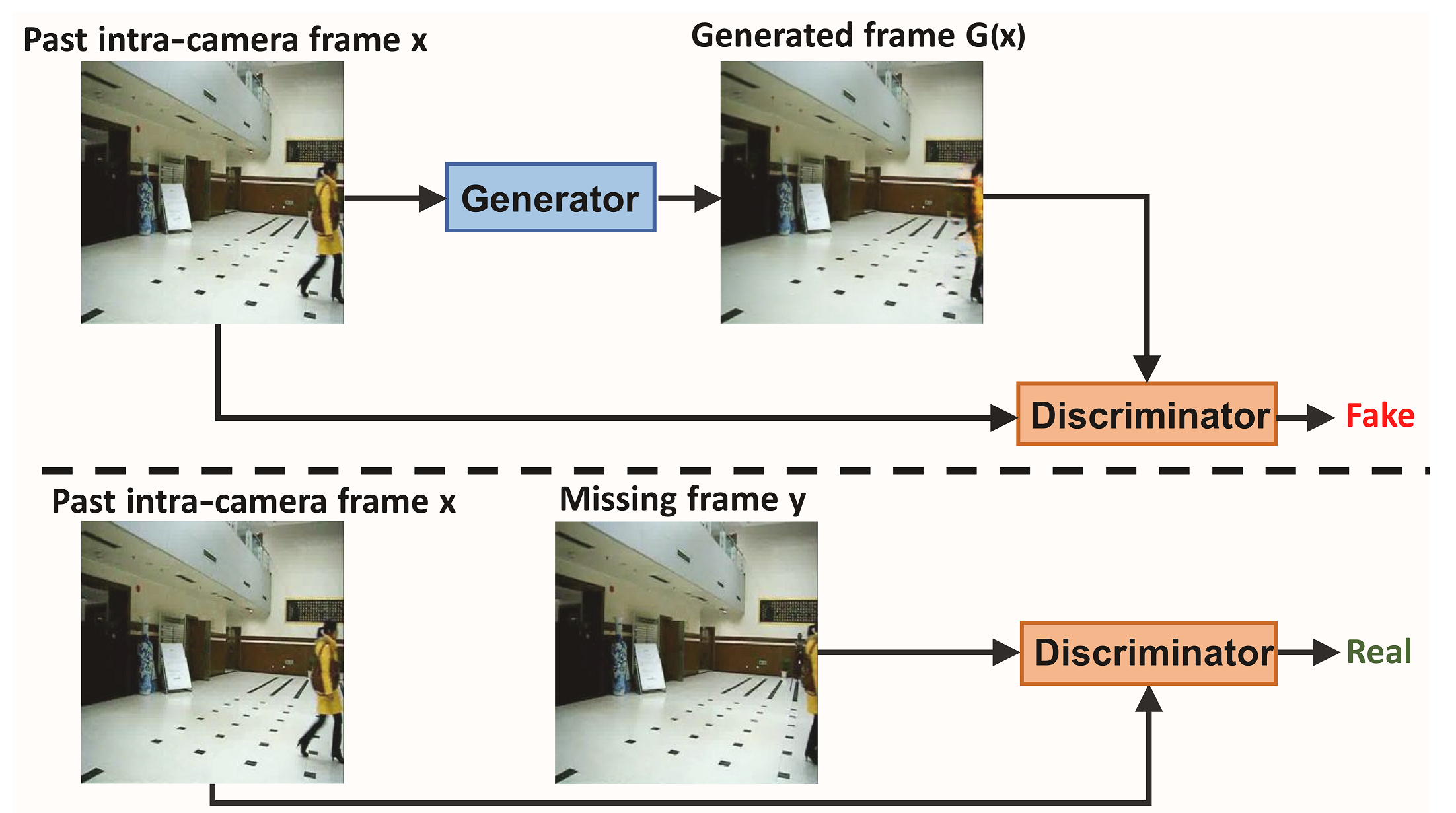}
	\caption{A training instance for the conditional GAN where the discriminator learns to classify between fake and real  frames and the generator learns to fool the discriminator.}
	\label{fig:fig4}
\end{figure}

In conditional GANs, a mapping is learned from an observed image $x$ and random noise vector $z$, to an output image $y$, $G : {x,z} \rightarrow y$ where the generator $G$ learns to generate outputs close to real images indistinguishable by the discriminator $D$ \cite{isola2017image}. The discriminator $D$ learns to efficiently detect the fake outputs generated by $G$. The objective function of the conditional GAN is as follows:
\begin{align}\label{eq:1}
G^* &= E_{x,y} [\log D(x,y)] +	E_{x,z}[\log (1-D(x,G(x,z))] \nonumber \\
&~~~~~~~~~~~~~~~~~~~~~+ \lambda E_{x,y,z}[\lVert y-G(x,z) \rVert_1]
\end{align}
Here, $E_{x,y,z}[\lVert y-G(x,z) \rVert_1]$ is the $L1$ loss to reduce blurring. Let us assume that there are $n$ overlapping cameras available in a multi-camera scenario. The $i^{th}$ frame, $C_t^i$, is missing in the target camera. First, we generate two representations of the missing frame from the past and future frame within the camera using two separate conditional GANs. We generate $(\hat{C}_{t}^i|{C}_{t}^{i-k})$ using the past $(i-k)^{th}$ frame and $(\hat{C}_{t}^i|{C}_{t}^{i+k})$ using the future $(i+k)^{th}$ frame. In our case, $k$ can be any arbitrary number based on availability. We generate different representations of the missing frame from the corresponding frame in other reference cameras i.e., generate $(\hat{C}_{t}^i|{C}_{r_{j}}^i)$  where $j = 1\dots n$. Basically the network learns a mapping from the observed frames ($C_t^{i-k}$, $C_t^{i+k}$, and $C_{r_j}^i$) to the missing frame $C_t^i$. In accordance with \eqref{eq:1}, $C_t^{i-k}$, $C_t^{i+k}$, and $C_{r_j}^i$ are $x$ and $C_t^i$ is $y$. A training instance is shown in Fig.~\ref{fig:fig4}.

The generated frame tries to resemble the missing frame in terms of the $L1$ loss along with fooling the discriminator. Following \cite{goodfellow2014generative}, we alternate between a gradient descent step upon $D$ and one upon $G$. Also, in accordance with \cite{goodfellow2014generative}, the training maximizes $\log D(x, G(x, z))$. We divide the objective function in \eqref{eq:1} by $2$ during optimizing $D$ to slow down it learning rate relative to $G$. To optimize the network, we use a minibatch stochastic gradient descent with an adaptive sub-gradient method (Adam) \cite{kingma2014adam} and a learning rate of $0.0002$.

During testing, we merge all the generated frames using a weighted average. The weights are chosen by maximizing the average PSNR on a smaller validation set. The more adjacent the available frames are in the target camera, the more weight is given to the representations learned from them than those from the reference cameras. Please note that, since the cameras are partially overlapped, we incorporate the multi-view representation only when there is a person/object present in the overlapping zone. 


\begin{figure*}[t!]
	\centering
	\includegraphics[width=0.7\linewidth] {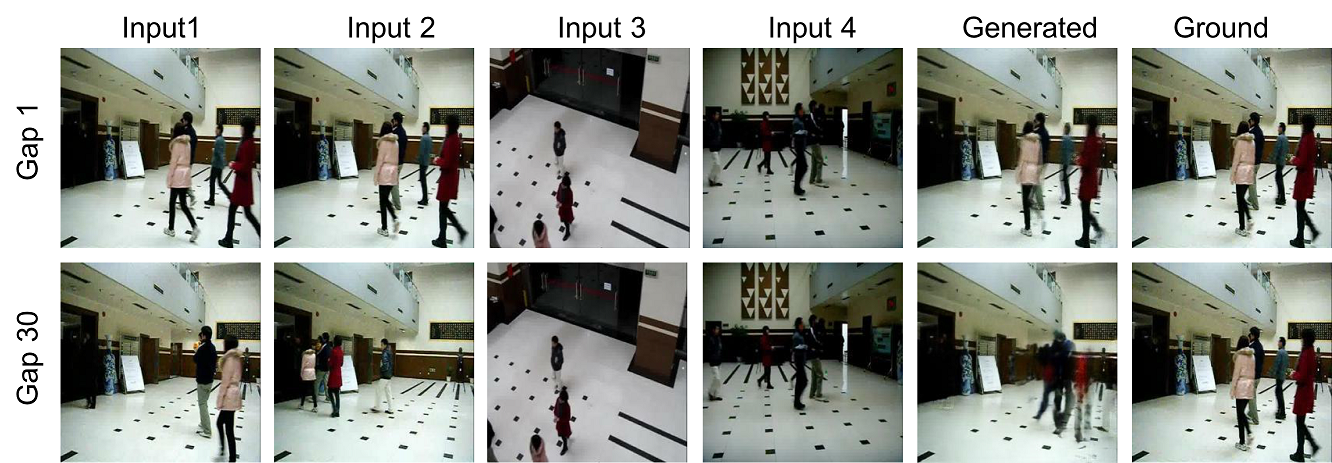}
	\caption{Two examples from Office Lobby Dataset where Input 1, Input 2, Input 3, and Input 4 are the preceding and the following frames of camera $1$, and the correspoinding frames of camera $2$ and $3$ respectively. As we increase the gap between the preceding and following frames with the missing frame, frames of camera $2$ and camera $3$ become more important. For example, due to the large number of missing frames in gap $30$, the women in red dress is not visible yet in input 1 and her position is far away in input $2$. Still, a person wearing a red dress is visible in the correct position of the generated frame incorporating information from the other two cameras.}
	\vspace{-3mm}
	\label{fig:fig7}
\end{figure*}

\section{EXPERIMENTS} \label{section:Exp}

\subsection{Dataset and Preprocessing}


\noindent{\bf Office Lobby Dataset.} Office Lobby Dataset is a multi-view summarization dataset where $3$ video clips are captured by $3$ cameras \cite{fu2010multi}. The cameras are not completely overlapping and the videos have different brightness levels across multi-views. The approximate offset between camera $1$ and $2$ is about $4.1s$ and between camera $1$ and $3$ is about $1.33s$. To make an approximate synchronization of the inter-camera frames, these offset values were taken into account while extracting and aligning the frames from different cameras. 

\noindent{\bf KTH Human Action Dataset.} KTH Human Action Dataset consists of $6$ types of human activities (boxing, handclapping, handwaving, jogging, running, and walking). These actions are performed by $25$ subjects in four different scenarios: outdoors, outdoors with scale variation, outdoors with different clothes, and indoors with lighting variation \cite{schuldt2004recognizing}.

\subsection{Results}
\noindent{\bf Objective.}
The main objective of these experiments is to evaluate the quality of the reconstructed frames in multi-camera scenario. We show how the overlapping cameras become more and more important as the distance is increased between the intra-camera frames and the missing frame.  

\noindent{\bf Performance Measure.}
The evaluation metrics we use are PSNR (Peak Signal-to-Noise Ratio) and SSIM (Structural Similarity Index). SSIM estimates how structurally close the reconstructed frame is to the original one. For both of these metrics, higher value indicates better performance. There is no existing work on multi-view frame reconstruction to compare our method with. To show the effectiveness of our method in a single camera scenario, we compare with a state-of-the-art reconstruction method \cite{sun2018temporally}.


\noindent{\bf Experimental Setup.}
We use the standard $80:20$ split for training and testing and use TensorFlow \cite{abadi2016tensorflow} to train our network on a NVIDIA Tesla K80 GPU.

\noindent{\bf Quantitative Evaluation.}
Our reconstruction results on Office Lobby Dataset increasing the distance between the missing frame and the available intra-camera past and future frames (multiple frames missing) are shown in Table~\ref{tab:Table1}. We consider different lengths (gap) of missing frame 
while testing which are selected in a sliding window manner. Comparisons of our reconstruction results on KTH Human Action Dataset are shown in Table~\ref{tab:Table2}. We achieve comparable PSNR and SSIM with those reported in \cite{sun2018temporally}. 

\begin{table}[h!]
	\begin{center}
		\begin{tabular}{|@{\hspace{0.8\tabcolsep}}c@{\hspace{0.8\tabcolsep}}|@{\hspace{0.8\tabcolsep}}c@{\hspace{0.8\tabcolsep}}|@{\hspace{0.8\tabcolsep}}c@{\hspace{0.8\tabcolsep}}|@{\hspace{0.8\tabcolsep}}c@{\hspace{0.8\tabcolsep}}|@{\hspace{0.8\tabcolsep}}c@{\hspace{0.8\tabcolsep}}|@{\hspace{0.8\tabcolsep}}c@{\hspace{0.8\tabcolsep}}|@{\hspace{0.8\tabcolsep}}c@{\hspace{0.8\tabcolsep}}|}
			\hline 
			Gap & \multirow{2}{*}{1}& \multirow{2}{*}{3} & \multirow{2}{*}{5} & \multirow{2}{*}{7} & \multirow{2}{*}{15} & \multirow{2}{*}{30}\\
			(frames) &  &  &  &  &  & \\
			\hline \hline
			PSNR & 32.06 & 29.28 & 28.10 & 27.19 & 25.56 & 25.17 \\
			SSIM  & 0.95 &0.92 &0.91 &0.90 &0.88 &0.87 \\
			\hline
		\end{tabular}
	\end{center}
	\vspace{-3mm}
	\caption{Multi-view Reconstuction Performance for Office Lobby Dataset.}
		\label{tab:Table1}
\end{table}
\begin{table} [h!]
	\begin{center}
		\begin{tabular}{|c|c|c|}
			\hline
			Method &PSNR & SSIM\\
			\hline \hline
			Proposed Method& 35.03& 0.93 \\
			LSTM-Based Method \cite{sun2018temporally}& 35.40& 0.96 \\
			\hline
		\end{tabular}
	\end{center}
	\vspace{-3mm}
	\caption{Single-view Reconstuction Performance Comparisons for KTH Human Action Dataset.}
	\label{tab:Table2}
\end{table}

\noindent{\bf Qualitative Evaluation.}
Some example reconstructed frames with the conditional input frames and the ground truth missing frames are shown in Fig.~\ref{fig:fig7}.


\noindent{\bf Ablation Study.}
The comparison of achieved PSNR using only the intra-camera view of camera $1$ vs. using multi-view reconstruction in Office Lobby Dataset is shown in Table~\ref{tab:Table3} as ablation study which justifies the integration of multi-view specially when the gap is large between the missing frame and the available intra-camera frames.

\begin{table}[h!]
	\begin{center}
		\begin{tabular}{|@{\hspace{0.8\tabcolsep}}c@{\hspace{0.8\tabcolsep}}|@{\hspace{0.8\tabcolsep}}c@{\hspace{0.8\tabcolsep}}|@{\hspace{0.8\tabcolsep}}c@{\hspace{0.8\tabcolsep}}|@{\hspace{0.8\tabcolsep}}c@{\hspace{0.8\tabcolsep}}|@{\hspace{0.8\tabcolsep}}c@{\hspace{0.8\tabcolsep}}|@{\hspace{0.8\tabcolsep}}c@{\hspace{0.8\tabcolsep}}|@{\hspace{0.8\tabcolsep}}c@{\hspace{0.8\tabcolsep}}|}
			\hline 
			Gap & \multirow{2}{*}{1}& \multirow{2}{*}{3} & \multirow{2}{*}{5} & \multirow{2}{*}{7} & \multirow{2}{*}{15} & \multirow{2}{*}{30}\\
			(frames) &  &  &  &  &  & \\
			\hline \hline
			Single & 32.06 & 29.24 & 28.02 & 27.02 & 24.17 & 23.97 \\
			Multi  & 32.06 & 29.28 & 28.10 & 27.19 & 25.56 & 25.17 \\
			\hline
		\end{tabular}
	\end{center}
	\vspace{-3mm}
	\caption{Ablation Study for Frame Reconstruction in Office Lobby Dataset considering Single-View vs. Multi-View.}
		\label{tab:Table3}
\end{table}

\section{Conclusions} \label{section:Con}
In this work, we proposed an adversarial learning framework for frame reconstruction in multi-camera scenario when one or more frames are missing. We learned the representation of the missing frame conditioned on the past and future frames within that camera as well as the corresponding frames in other overlapping cameras using conditional GAN and merged them together using a weighted average. 

\section{Acknowledgements} \label{section:acknowledgements}
This work was partially supported by NSF grant 1544969 from the Cyber-Physical Systems program.

\bibliographystyle{IEEEbib}
\bibliography{mybib_acm}

\end{document}